\documentclass[10pt]{article} % For LaTeX2e
\usepackage[preprint]{tmlr}
\usepackage{amssymb}
\usepackage{amsmath}
\usepackage{amsthm}

\usepackage[utf8]{inputenc} % allow utf-8 input
\usepackage[T1]{fontenc}    % use 8-bit T1 fonts
\usepackage{hyperref}       % hyperlinks
\usepackage{url}            % simple URL typesetting
\usepackage{booktabs}       % professional-quality tables      
\usepackage{bm}
\usepackage{nicefrac}       % compact symbols for 1/2, etc.
\usepackage{microtype}      % microtypography
\usepackage{amsmath, amsthm, amssymb, amsmath, amsfonts}
\usepackage{mathabx} %for widebar
\usepackage{physics}
\usepackage{multirow}
\usepackage{graphicx}
\usepackage[belowskip=-6pt]{caption}
\usepackage{subcaption}
\usepackage{xcolor}
\usepackage{algorithm}
\usepackage{algcompatible}
\usepackage{lineno}
\usepackage{subcaption}
\graphicspath{ {./images/} }
\newcommand{\mD}{\mathcal{D}}
\newcommand{\bI}{\mathbf{I}}
\newcommand{\mI}{\mathcal{I}}
\newcommand{\bK}{\mathbf{K}}
\newcommand{\bell}{\bm\ell}
\newcommand{\mL}{\mathcal{L}}
\newcommand{\mvec}{\mathbf{m}}
\newcommand{\bq}{\mathbf{q}}
\newcommand{\bx}{\mathbf{x}}
\newcommand{\bX}{\mathbf{X}}
\newcommand{\by}{\mathbf{y}}
\newcommand{\R}{\mathbb{R}}
\newcommand{\abss}[1]{| #1 |}

\newcommand{\zero}{\mathbf{0}}
\newcommand{\argmax}{\arg\!\max}
\newtheorem{remark}{Remark}
\newenvironment{keywords}
  {\vspace{1em}\noindent\textbf{Keywords:}\slshape}
  {\par\vspace{1em}}
\title{Label Propagation for Physics-Informed Neural Networks and Physics-Informed Gaussian Processes}

\author{\name Ming Zhong \email mzhong3@central.uh.edu \\
      \addr Department of Mathematics\\
      University of Houston
      \AND
      \name Dehao Liu \email dehaoliu@binghamton.edu \\
      \addr Department of Mechanical Engineering \\
      Binghamton University
      \AND
      \name Raymundo Arroyave \email rarroyave@tamu.edu\\
      \addr Department of Material Science $\&$ Engineering \\
      Texas A$\&$M University
      \AND
      \name Ulisses Braga-Neto \email ulisses@tamu.edu \\
      \addr Department of Electrical $\&$ Electronic Engineering\\
      Texas A$\&$M University}
\begin{document}
\maketitle
\begin{abstract}
We present a series of empirical results of the application of semi-supervised label propagation techniques in training physics-informed machine learning methods. This includes self-training of physics-informed neural networks and physics-informed Gaussian processes in isolation, and the integration of the two via co-training, therefore establishing a hybrid between these two main classes of physics-informed machine learning. We demonstrate via extensive numerical experiments how these methods can ameliorate the issue of propagating information from boundaries into the physical domain, including information from initial conditions in the case of solving stiff time-dependent partial differential equations, which is known to be a common failure mode of physics-informed machine learning. 
\end{abstract}
%% Keywords
\section{Introduction}\label{sec:intro}
Typical data-driven Machine Learning methodologies do not incorporate physical understanding of the problem domain. Furthermore, in many scientific domains, high-fidelity data are expensive or time consuming to obtain. Physics-informed machine learning is an emerging area that addresses both of these problems by introducing regularizing constraints obtained from physical laws, allowing prediction of the future performance of complex systems using sparse data \citep{yang2019physics,chen2021solving,raissi2018hidden,raissi2019physics,liu2019multi,liu2021dual,jin2021nsfnets,cai2021deepm,rad2020theory,zhu2021machine}. However, successfully training physics-informed machine learning models is difficult \citep{wang2020understanding,wang2020and,mcclenny2020self, coutinho2022physics}. One of the basic issues is the difficulty in propagating information away from the data in order to make extrapolating predictions, especially in the case of stiff PDEs. 

In the literature on semi-supervised regression, self-training consists of progressively attributing labels to unlabeled points with high confidence of prediction during training \citep{yarowsky1995unsupervised,rosenberg2005semi}. Similarly, the co-training paradigm in semi-supervised learning is based on the observation that the disagreement among independent predictors limits their individual prediction errors \citep{kunselman2020semi}. In co-training, one classifier or regressor makes predictions on the unlabeled data used by another classifier or regressor, respectively, and the process is iterated to convergence \citep{zhou2005semi,brefeld2006efficient}. 

In this paper, we follow these ideas and develop semi-supervised self-training and co-training algorithms to train physics-informed neural networks (PINNs) and physics-informed Gaussian processes (PIGPs). In other words, PINNs and PIGPs can be trained in isolation (self-training) or simultaneously to achieve a PINN-PIGP hybrid. In particular, this allows uncertainty quantification in the predictions made by the PINN, via the posterior variance of its co-trained PIGP.

Our key insight is that collocation-based machine learning solvers used to train PINNs and PIGPs are a form of specialized semi-supervised regression, where the initial condition and Dirichlet boundary condition data, and possibly additional observed data points (with varying levels of observation noise) within the computational domain play the role of labeled data, and collocation points play the role of unlabeled data. Thus, self-training schemes can be applied by establishing a fidelity criterion for predictions on the collocation points and progressively adding them to the set of ``labeled'' data points. This can be done on a tentative basis, where points can be added and removed from the labeled set as training progresses. In the case of PIGPs, the fidelity criterion is readily available, consisting of the variance of the posterior distribution. In the case of PINNs, the criterion can be the proximity to currently labeled data and the value of the PDE residual at the collocation point. Self-adaptive weights can also be used and updated based on this self-training scheme. In addition to self-training, PINNs and PIGPs can be co-trained by having one make reliable predictions on the unlabeled points used by the other. As in self-training, this requires a criterion of prediction fidelity to decide when a point should be labeled. As such, we are able to use the fidelity metrics developed in self-training in co-training. Co-training has the added benefit of having the PIGP provide reliable uncertainty quantification metrics for the predictions made by the co-trained PINN.
\subsection{Related Work}
The baseline PINN algorithm, though remarkably successful in many applications, can produce inaccurate approximations or fail to converge entirely, in the solution of certain ``stiff'' PDEs. There is a growing literature of evidence that this occurs due to an imbalance between the data-fitting and residual components of the PINN loss function \citep{wight2020solving,shin2020convergence,wang2020understanding,wang2020and}. Gradient descent will drive the residual loss to zero faster than the data-fitting component, which prevents convergence to the correct solution. In forward problems, this problem may be especially acute, since all the data are confined to initial and boundary conditions, and there may not be any data inside the domain of the PDE. For example, in time-evolution problems, it may be difficult for a neural network during training to advance information in the initial condition to later times, as has been observed by several authors, e.g.~\cite{wight2020solving,mcclenny2020self,krishnapriyan2021characterizing,wang2022respecting,haitsiukevich2022improved}. In~\cite{wight2020solving}, an approach to propagate information forward in time was proposed, whereby the time axis is divided into several smaller intervals and PINNs are trained separately on them, sequentially, beginning with the interval close to the initial condition; this approach is time consuming due to the need to train multiple PINNs. It was pointed out in both \cite{wight2020solving} and \cite{mcclenny2020self} that it is necessary to weight the initial condition data and residual points near $t=0$ in order to propagate the information forward in a well-known nonlinear Allen-Cahn PDE benchmark. In the case of the self-adaptive PINNs in \cite{mcclenny2020self}, this fact is automatically discovered during training, and again in a Wave equation benchmark with a complex initial condition. The same theme appears in \cite{krishnapriyan2021characterizing}, where a sequential progression of curriculum-like training is proposed. A time-marching scheme in which the PINN learns at each time step sequentially is proposed for solving Allen--Cahn-like PDEs~\citep{krishnapriyan2021characterizing, lwight2021}. In \cite{wang2022respecting}, the authors attribute the problem to a lack of ``causality'' in standard PINN training and propose to correct this by a weighting scheme that progressively weights the data from early to later times, using the information from the PDE residual itself, in a manner that is reminiscent of the self-adaptive weighting in \cite{mcclenny2020self}.  Another way of enforcing ``causality'' is to use a hard-constrained transformation as in~\cite{bhao2026}. In~\cite{haitsiukevich2022improved}, the authors use an ensemble of homogeneous PINNs to estimate prediction confidence and progressively propagate labels, whereas our approach co-trains structurally independent PINN and PIGP models that exchange reliable pseudo-labels based on their distinct confidence criteria.

In semi-supervised learning, a predictor is trained using both labeled and unlabeled data~\citep{zhu2005semi}. Self-training is a semi-supervised learning technique that fits a predictor progressively, by assigning ``pseudo-labels'' to data that is initially unlabeled, and adding these data to the labeled data; unlabeled data points receive a pseudo-label once the prediction of their label becomes sufficiently confident \citep{rosenberg2005semi}. Some pseudo-labeled points could later in training become unlabeled again, if their confidence drops below a certain threshold. How to attribute confidence to a pseudo-labeled point is therefore critical to the performance of the final trained predictor; criteria used for this include proximity to already labeled data by means of a distance function or graph neighborhood structure. On the other hand, co-training is a semi-supervised technique where two or more predictors are trained simultaneously, and their combined decisions are used to assign pseudo-labels, which are used to train all predictors~\citep{blum1998combining}. For example, the feature set may be partitioned into two smaller sets. A classifier or regressor may be trained to make predictions on the unlabeled data used by another classifier or regressor, respectively, and the process is iterated until convergence \citep{zhou2005semi,brefeld2006efficient}. 

In typical machine learning literature, a hybrid method combining a neural network regressor and a Gaussian process regressor has been a popular research topic.  In~\cite{pmlr-v51-wilson16}, a GP-NN hybrid uses an NN to learn the kernel and a GP on the prediction side. A closely related and slightly earlier formulation can be found in~\cite{pub13786}.  Another similar idea is called Neural Processes~\citep{pmlr-v80-garnelo18a}, which involves designing neural networks from the ground up to have GP-like capabilities.  Our approach is different from these previous methods: for one thing, it has the physics-informed training component; for another, the PINN and PIGP are independent from each other (structure-wise) and interact through the training process.
\subsection{Contributions of this Work}
In this paper, we propose self-training and co-training algorithms for training PINNs and PIGPs separately or simultaneously. Our contributions can be detailed as follows.
\begin{itemize}
\item[$(I)$] We develop self-trained and co-trained PINN and PIGP algorithms, and investigate their stability, training speed, and prediction accuracy.
\item[$(II)$] We investigate how self-training and co-training ameliorate the issue of propagating information forward in time, which is a common failure mode of PINNs.
\item[$(III)$] We propose the first PINN-PIGP hybrid method in the literature, namely by co-training PINNs and PIGPs; moreover, it can also combine a PINN and GPR, or an NN and PIGP. As in self-training, this requires a criterion of prediction fidelity to decide when a point should be labeled. As such, we will be able to use the fidelity metrics developed in self-training in co-training. Co-training has the added benefit of having the PIGP provide reliable UQ metrics for the predictions made by the co-trained PINN.
\end{itemize}
\section{Methodology}
Before we introduce the enhanced training framework for physics-informed neural networks (PINNs) for solving Partial Differential Equations (PDEs), we need to lay down the basic notation about PDEs.  Let $u$ be an unknown scalar function defined on $[0, T]\times\Omega$ with $\Omega \subset \R^d$ ($d \ge 1$) with a Lipschitz boundary $\partial\Omega$.  Moreover, we take $\widebar\Omega = \Omega\cup\partial\Omega$. We assume that $u$ solves a PDE if the following equations are satisfied
\begin{equation}\label{eq:pde_setup}
\begin{cases}
u_t + \mathcal{P}(u)(t, \bx) &= f(t, \bx), \quad (t, \bx) \in (0, T]\times\Omega, \\
u(0, \bx) &= u_0(\bx), \quad \bx \in \widebar\Omega, \\
\mathcal{B}(u)(t, \bx) &= g(t, \bx), \quad (t, \bx) \in [0, T]\times\partial\Omega, \\
\end{cases}
\end{equation}
where $\mathcal{P}$ and $\mathcal{B}$ are partial differential operators, and the two functions $f, g$ are assumed to have sufficient regularity conditions and compatibility conditions (e.g., $u_0(\bx) = g(0, \bx)$ for $\bx \in \partial\Omega$ in the Dirichlet case) so that the existence and uniqueness of solutions for the aforementioned PDE are guaranteed.  
\begin{remark}
Although our definition of PDEs seems to focus on time-dependent PDEs, for those PDEs without the time variable, we simply take $u_t = 0$ and drop the terms related to time in our loss and focus only on the spatial variable $\bx \in \Omega$.
\end{remark}
For example, for heat equations, i.e. $u_t = au_{xx}$, with an initial condition $u(0, x) = u_0(x)$ for $x \in [a, b]$ and a Dirichlet boundary condition, $u(t, a) = u(t, b) = 0$ for $t \in [0, T]$; we will let $\mathcal{P} = -a\partial_{xx}$, $\mathcal{B} = I_{id}$ (the identity operator) for $x = a$ or $b$, $f(t, x) = 0$, and $g(x) = 0$.

Solving \eqref{eq:pde_setup} by a PINN involves searching for an approximate solution $u_{NN}$ in a space of deep neural networks $\mathcal{H}_{\text{NN}}$ (of the same depth, same number of neurons on each hidden layer, and same activation function), which is a minimizer of the following loss functional
\begin{equation}\label{eq:loss}
\begin{aligned}
\text{Loss}(u_{NN}) &= \frac{1}{N_{CL}}\sum_{i = 1}^{N_{CL}}\abss{(u_{NN})_t(t_i^{CL}, \bx_i^{CL}) + \mathcal{P}(u_{NN})(t_i^{CL}, \bx_i^{CL}) - f(t_i^{CL}, \bx_i^{CL})}^2 \\
&\quad + \frac{1}{N_{IC}}\sum_{i = 1}^{N_{IC}}\abss{u_{NN}(0, \bx_i^{IC}) - u_0(\bx_i^{IC})}^2 \\
&\quad + \frac{1}{N_{BC}}\sum_{i = 1}^{N_{BC}}\abss{\mathcal{B}(u_{NN})(t_i^{BC}, \bx_i^{BC}) - g(t_i^{BC}, \bx_i^{BC})}^2,
\end{aligned}
\end{equation}
where the set $\{(t^{CL}_i, \bx_i^{CL})\}_{i=1}^{N_{CL}} \subset (0, T]\times\Omega$ contains the collocation points, $\{(t_i^{IC}, \bx_i^{IC})\}_{i=1}^{N_{IC}} \subset \{0\}\times\widebar\Omega$ the initial condition points, and $\{(t_i^{BC}, \bx_i^{BC})\}_{i=1}^{N_{BC}} \subset [0, T]\times\partial\Omega$ the boundary condition points, respectively.   

\textbf{Gaussian Process Regression Framework}: In order to proceed with the setup for Physics Informed Gaussian Processes, we need to introduce the Gaussian Process regression (GPR) first.  Consider the set of observation locations $\{\bx_1, \bx_2, \cdots, \bx_M\}$ with each $\bx_m \in \R^{d}$ (here we are considering the points as a time-space unit, i.e. $\bx = (t, x_1, \cdots, x_{d - 1})$), and the observed values at these locations are $\by = \begin{bmatrix} y_1 & y_2 & \cdots & y_M \end{bmatrix}^\top \in \R^M$.  The GPR method assumes that the observation $\by$ is a realization of the following $M$-dimensional random vector that satisfies a multivariate Gaussian distribution, where
\[
\by = \begin{bmatrix} u(\bx_1) \\ u(\bx_2) \\ \vdots \\ u(\bx_M)\end{bmatrix}, \quad \text{and} \quad
\text{Pr}(\by \big| \bX) = \mathcal{N}(\by \big| \mvec, \bK), \quad \bX = \begin{bmatrix}\bx_1^\top \\ \bx_2^\top \\ \vdots \\ \bx_M^\top \end{bmatrix},
\]
with
\[
\mvec = \begin{bmatrix} \mu(\bx_1) \\ \mu(\bx_2) \\ \vdots \\ \mu(\bx_M)\end{bmatrix} \quad \text{and} \quad \bK = [K_{i, j}] = [k(\bx_i, \bx_j)], 
\]
where $\mu: \R^d \rightarrow \R$ is a mean function and $k: \R^d\times\R^d \rightarrow \R$ is a kernel function.  Here we assume $\by$ to have zero mean, i.e. $\mvec = \zero$. The similarity between $\bx_i$ and $\bx_j$ is given by the kernel function $k$.  Then when a new data point $\bx_*$ comes in, the joint distribution of $y_* = u(\bx_*)$ with $\by$ is given as
\[
\begin{bmatrix} y_* \\ \by \end{bmatrix} \sim \mathcal{N}\big(\begin{bmatrix} \mu(\bx_*) \\ \mvec \end{bmatrix}, \begin{bmatrix} \bK_{**} & \bK_* \\ \bK_*^\top & \bK \end{bmatrix}\big)
\]
where $\bK_{**} = K(\bx_*, \bx_*)$, $\bK_{*} = K(\bx_*, \bX)$ and $\bK = K(\bX, \bX)$, again we assume $\mu(\bx_*) = 0$.  With this joint distribution, we can find $\text{Pr}(y_*\big| \by, \bX, \bx_*)$ using the following
\[
\text{Pr}(y_*\big| \by, \bX, \bx_*) = \mathcal{N}(y_* \big| \bK_*\bK^{-1}\by, \bK_{**} - \bK_*\bK^{-1}\bK_*^\top).
\]
When the observation data are corrupted by noise, i.e., $y_m = u(\bx_m) + \epsilon_m$, where the $\epsilon_m$ are i.i.d. additive Gaussian noise terms with zero mean and variance $\sigma_u^2$, then
\[
\begin{bmatrix} y_* \\ \by \end{bmatrix} \sim \mathcal{N}\big(\begin{bmatrix} \mu(\bx_*) \\ \mvec \end{bmatrix}, \begin{bmatrix} \bK_{**} & \bK_* \\ \bK_*^\top & \bK + \sigma_u^2\bI \end{bmatrix}\big)
\]
Therefore the updated conditional probability with the noise is
\[
\text{Pr}(y_*\big| \by, \bX, \bx_*) = \mathcal{N}(y_* \big| \bK_*(\bK + \sigma_u^2\bI)^{-1}\by, \bK_{**} - \bK_*(\bK + \sigma_u^2\bI)^{-1}\bK_*^\top).
\]
The shift of $\bK$ by $\sigma_u^2\bI$ might improve computing $(\bK + \sigma_u^2\bI)^{-1}$ when $\sigma_u$ is small.  However, the whole regression depends on finding a suitable kernel function $k$; hence, we need to introduce a method to find the optimal kernel by the log-likelihood of the marginal, i.e.,
\[
\log(\text{Pr}(\by\big| \bX)) = -\frac{1}{2}\by^\top(\bK + \sigma_u^2\bI)^{-1}\by - \frac{1}{2}\log(\det(\bK + \sigma_u^2\bI)) - \frac{M}{2}\log(2\pi).
\]
If $k$ is parametrized as follows
\[
k(\bx_i, \bx_j) = \ell_1\exp(-\frac{1}{2\ell_2}(\bx_i - \bx_j)^\top(\bx_i - \bx_j)),
\]
where the parameter $\ell_1$ determines the vertical span of the function and $\ell_2$ controls the rate at which the correlation between two points decreases with increasing distance.  Let $\bell = \begin{bmatrix} \ell_1 & \ell_2 \end{bmatrix}^\top$. The optimal parameter $\bell$ is chosen to maximize the log-marginal likelihood
\[
\bell_* = \argmax_{\bell}\log(\text{Pr}(\by \big| \bX, \bell)).
\]

Next, we consider any general linear PDE as $\mL(u)(\bx) = f(\bx)$, for $\bx \in \mD \subset \R^d$. Under the setting of a time-dependent PDE, i.e. 
\[
\begin{aligned}
    u_t(t, x) + \mathcal{P}(u)(t, x) = h(t, x) &\Rightarrow \mL = \partial_t + \mathcal{P}, \quad f(\bx) = h(\bx) \quad, \bx = (t, x) \in (0, T]\times\Omega, \\
    u(0, x) = u_0(x) &\Rightarrow \mL = \mI, \quad f(\bx) = u_0(x), \quad \bx \in \{0\}\times\bar\Omega, \\
    u(t, x) = g(t, x) &\Rightarrow \mL = \mI, \quad f(\bx) = g(\bx), \quad \bx \in [0, T]\times\partial\Omega,
\end{aligned}
\]
Here $\mI$ is the identity operator and the domain $\mD = [0, T]\times\bar\Omega$ with $\bar\Omega = \Omega\cup\partial\Omega$.  When we have initial conditions and/or Dirichlet type of boundary conditions, we actually call them direct observations of the PDE solution $u$ (but on a limited set, i.e. IC and BC points).  Given such a linear PDE problem, we can break the noisy observation into two parts, i.e. $\{\bX_u, \by_u\}$ and $\{\bX_f, \by_f\}$ for $u(\bx)$ (direct observation, labeled data) and $f(\bx)$ (indirect observation, prior knowledge from the PDE).  Then if $u$ is modeled as a Gaussian process, with $0$ mean and a covariance function $k_{uu}$ as follows
\[
u \sim \mathcal{GP}(0, k_{uu}).
\]
Since any linear transformation of a Gaussian process is still a Gaussian process, we have
\[
f = \mL(u) \sim \mathcal{GP}(0, k_{ff}), \quad k_{ff}(\bx, \bx') = \mL_{\bx}\mL_{\bx'}k_{uu}(\bx, \bx').
\]
The covariance between $u(\bx)$ and $f(\bx')$ is given by $k_{uf}(\bx, \bx') = \mL_{\bx'}k_{uu}(\bx, \bx')$; similarly, the covariance between $f(\bx)$ and $u(\bx')$ is $k_{fu}(\bx, \bx') = \mL_{\bx}k_{uu}(\bx, \bx')$.  Define the noisy observation vectors by 
\[
\by = \begin{bmatrix} \by_u \\ \by_f \end{bmatrix}, \quad \by_u = \begin{bmatrix} u(\bx_u^1) \\ \vdots \\ u(\bx_u^{M_u})\end{bmatrix} + \boldsymbol{\epsilon}_u, \quad \by_f = \begin{bmatrix} f(\bx_f^1) \\ \vdots \\ f(\bx_f^{M_f})\end{bmatrix} + \boldsymbol{\epsilon}_f,
\]
where $\boldsymbol{\epsilon}_u \sim \mathcal{N}(\zero, \sigma_u^2\bI)$ and $\boldsymbol{\epsilon}_f \sim \mathcal{N}(\zero, \sigma_f^2\bI)$ are independent of each other and of the latent process. Then the joint distribution is
\[
\by \big| \bX_u, \bX_f \sim \mathcal{N}(\zero, \bK), \quad \bK = \begin{bmatrix}k_{uu}(\bX_u, \bX_u) + \sigma_u^2\bI & k_{uf}(\bX_u, \bX_f) \\ k_{fu}(\bX_f, \bX_u) & k_{ff}(\bX_f, \bX_f) + \sigma_f^2\bI \end{bmatrix},
\]
where $\sigma_u^2$ and $\sigma_f^2$ are the variances of the observation noise to $u$ and $f$, and
\[
\bX_u = \begin{bmatrix} (\bx_u^1)^\top \\ \vdots \\ (\bx_u^{M_u})^\top \end{bmatrix} \quad \text{and} \quad \bX_f = \begin{bmatrix} (\bx_f^1)^\top \\ \vdots \\ (\bx_f^{M_f})^\top \end{bmatrix}
\]
For a new input point $\bx_*$, the predicted value of $u(\bx_*) = u_*$ is given by the following posterior distribution, 
\[
u_* \big| \by \sim \mathcal{N}(\bar{u}(\bx_*), k_u(\bx_*))
\]
where the mean and variance are
\[
\bar{u}(\bx_*) = \bq_u^\top\bK^{-1}\by, \quad k_u(\bx_*) = k_{uu}(\bx_*, \bx_*) - \bq_u^\top\bK^{-1}\bq_u, \quad \bq_u = \begin{bmatrix} k_{uu}(\bx_*, \bX_u)^\top \\ k_{uf}(\bx_*, \bX_f)^\top \end{bmatrix}.
\]
Similarly for $f(\bx_*) = f_*$,
\[
f_* \big| \by \sim \mathcal{N}(\bar{f}(\bx_*), k_f(\bx_*))
\]
where the mean and variance are
\[
\bar{f}(\bx_*) = \bq_f^\top\bK^{-1}\by, \quad k_f(\bx_*) = k_{ff}(\bx_*, \bx_*) - \bq_f^\top\bK^{-1}\bq_f, \quad \bq_f = \begin{bmatrix} k_{fu}(\bx_*, \bX_u)^\top \\ k_{ff}(\bx_*, \bX_f)^\top \end{bmatrix}.
\]
If the kernel function $k_{uu}$ is again parametrized by a list of parameters $\bell$, then we find the optimal $\bell_*$ by minimizing the negative log marginal likelihood 
\[
-\log(\text{Pr}(\by \big| \bell)),
\]
where 
\[
\text{Pr}(\by \big| \bell) = \mathcal{N}(\by \big| \zero, \bK), \quad \bK = \begin{bmatrix} k_{uu}(\bX_u, \bX_u; \bell) + \sigma_u^2\bI_u & k_{uf}(\bX_u, \bX_f; \bell) \\ k_{fu}(\bX_f, \bX_u; \bell) & k_{ff}(\bX_f, \bX_f; \bell) + \sigma_f^2\bI_f\end{bmatrix}
\]
Hence
\[
\log(\text{Pr}(\by \big| \bell)) = -\frac{1}{2}\by^\top\bK^{-1}\by - \frac{1}{2}\log(\det(\bK)) - \frac{M_u + M_f}{2}\log(2\pi).
\]
The main computational bottleneck is caused by the computation of $\bK^{-1}$, which can be remedied by using a Cholesky decomposition of $\bK$ and solving the resulting triangular systems, since $\bK$ is symmetric positive definite.
% There is a more efficient way of computing $\bK^{-1}$ given the way $\bK$ is partitioned.  Recall
% \[
% \bK = \begin{bmatrix} k_{uu}(\bX_u, \bX_u; \bell) + \sigma_u^2\bI_u & k_{uf}(\bX_u, \bX_f; \bell) \\ k_{fu}(\bX_f, \bX_u; \bell) & k_{ff}(\bX_f, \bX_f; \bell) + \sigma_f^2\bI_f\end{bmatrix} = \begin{bmatrix} A & B \\ C & D \end{bmatrix}
% \]
% where $A$ is a $M_u \times M_u$ and $D$ is a $M_f \times M_f$ matrix, if $M_f < M_u$, then we find $\bK^{-1}$ using the following formula
% \[
% \bK^{-1} = \begin{bmatrix} A^{-1} + A^{-1}BXCA^{-1} & -A^{-1}BX \\ -XCA^{-1} & X\end{bmatrix}, \quad X = (D - CA^{-1}B)^{-1};
% \]
% if $M_f > M_u$, then
% \[
% \bK^{-1} = \begin{bmatrix} W & -WBD^{-1} \\ -D^{-1}CW & D^{-1} + D^{-1}CWBD^{-1} \end{bmatrix} \quad W = (A - BD^{-1}C)^{-1}.
% \]
% This is taken from \url{https://jepusto.com/posts/inverting-partitioned-matrices/}.
\begin{remark}
PIGP is best suited for linear problems (i.e., when the differential operator is linear, the posterior distribution can be completely characterized using a Gaussian distribution with closed-form (analytic) expressions for posterior means and covariances, albeit the computation of the kernel is still intensive), and PINN is best suited for nonlinear problems (as neural networks have been proven to have universal approximation properties in various settings).  There are PIGP applications for nonlinear problems, such as~\cite{chen2021solving}; however most of them focus on the kernel/approximation perspective of the posterior means without computable posterior covariances. Based on this observation, when we consider the hybridization of PINN and PIGP for the co-training schemes, we mainly use them for linear PDEs; for nonlinear PDEs, we consider the PINN/GPR combination.
\end{remark}
\subsection{Propagation of Labels}
The difficulty of the training process (i.e. finding the minimizer) of PINN or PIGP is caused by the fact that the collocation points within the domain are not associated with a direct (or noisy) function value.  Hence, they are considered unlabeled points.  If we can find those labeled points (associated with function values) in an approximate sense, it can help us to train better and converge faster for PINN and/or PIGP.

Our main idea is that if a point has a sufficiently small residual error from a PINN estimator or a sufficiently small variance from a PIGP estimator, and it is also close to points with true labels, i.e., initial-condition or Dirichlet boundary-condition points, then it can be assigned a pseudo-label (a computed function value with noise), with the function value from the corresponding estimator thought of as a noisy input.  And by progressively labeling points in this fashion, we would be able to add these pseudo-labeled (and noisy) data points to the estimator and enhance the training results.  

We present two different algorithms for training the physics-informed machine learning approximations: one is about self-training (one trains itself for label propagation) for propagating the labels presented in Algorithm \ref{alg:self_train}, and the other one is using co-training (two estimators train each other for label propagation) for propagating the labels in Algorithm \ref{alg:co_train}.  We assume that $u_{\text{prox}}$ is an approximation from either the PINN or the PIGP, $\mathcal{S}_{CL} = \{(\bx_i^{CL}, f(\bx_i^{CL}))\}_{i = 1}^{N_{CL}}$ (set of collocation points and their forcing values), $\mathcal{S}_{BC} = \{(\bx_i^{BC}, g(\bx_i^{BC}))\}_{i = 1}^{N_{BC}}$ (set of boundary points), $\mathcal{S}_{Test} = \{\bx_i^{Test} \in \Omega\}_{i = 1}^{N_{Test}}$ (test points for pseudo-labels), and $\mathcal{S}_{PL} = \{(\bx_i^{PL}, u_{\text{prox}}(\bx_i^{PL}))\}_{i = 1}^{N_{PL}}$ (pseudo-label points).  We also define $\mathcal{S}_{CL}^{\bx} = \{\bx_i^{CL}\}_{i = 1}^{N_{CL}}$ (containing only the points without function values), and similarly for $\mathcal{S}_{BC}^{\bx}$, $\mathcal{S}_{PL}^{\bx}$. 

There are a few changes one can make to Algorithm \ref{alg:self_train}. First, one can generate a huge $\mathcal{S}_{Test}$ before the loop starts, and keep it the same by removing the candidate points from $\mathcal{S}_{Test}$; second, within the loop, we can also test not only the points from $\mathcal{S}_{Test}$ but also from $\mathcal{S}_{PL}^{\bx}$, in order to keep the best possible points in $\mathcal{S}_{PL}$ for each iterate. 
 Next, for the co-training algorithm presented in Algorithm \ref{alg:co_train}, we need two estimators, $u_{NN}$ from PINN and $u_{GP}$ from PIGP.  We use the same notation for the set of training points; however, we use two different pseudo-label sets, one $\mathcal{S}_{PL}^{NN}$ for PINN, and the other $\mathcal{S}_{PL}^{GP}$ for PIGP.
 \begin{algorithm}[t]
\caption{Self-Training}\label{alg:self_train}
\begin{algorithmic}
\REQUIRE $\mathcal{P}$, $\mathcal{B}$, $\Omega$, $\partial\Omega$, $f$, $g$, $I_{\max}$, $\sigma_1$, $\sigma_2$, and $N_{Test}$
\ENSURE$\mathcal{P}(u) = f$ on $\Omega$ and $\mathcal{B}(u) = g$ on $\partial\Omega$
\STATE Initialize $iter = 0$, $\mathcal{S}_{CL}$, $\mathcal{S}_{BC}$, and $\mathcal{S}_{PL} = \emptyset$ 
% AUTHOR CHECK: With iter starting at 0, this loop permits I_max + 1 passes. Confirm whether I_max is a final index or a count; the original stopping rule is retained.
\WHILE{$iter \le I_{\max}$}
\STATE Train $u_{\text{prox}}$ for a fixed number of epochs on $\mathcal{S}_{CL}$, $\mathcal{S}_{BC}$ and $\mathcal{S}_{PL}$
\STATE Randomly generate $\mathcal{S}_{Test}^{iter}$
\STATE Find all possible $\bx_i^{Test, iter}\in \mathcal{S}_{Test}^{iter}$, such that
\[
\abss{\mathcal{P}(u_{\text{prox}})(\bx_i^{Test, iter}) - f(\bx_i^{Test, iter})} < \sigma_1, \quad \text{and} \quad \text{dist}(\bx_i^{Test, iter}, \mathcal{S}_{BC}^{\bx} \cup \mathcal{S}_{PL}^{\bx}) < \sigma_2,
\]
\STATE then $(\bx_i^{Test, iter}, u_{\text{prox}}(\bx_i^{Test, iter}))$ is moved to $\mathcal{S}_{PL}$.
\STATE If $u_{\text{prox}}$ is generated by PIGP, we use $\text{Var}(u_{\text{prox}})(\bx_i^{Test, iter}) < \sigma_1$ instead.
\IF{No such $\bx_i^{Test, iter}$ is found}
  \STATE break
\ELSE
  \STATE $iter = iter + 1$
\ENDIF
\ENDWHILE
\end{algorithmic}
\end{algorithm}
\begin{algorithm}[!h]
\caption{Co-Training}\label{alg:co_train}
\begin{algorithmic}
\REQUIRE $\mathcal{P}$, $\mathcal{B}$, $\Omega$, $\partial\Omega$, $f$, $g$, $I_{\max}$, $\sigma_1$, $\sigma_2$, and $N_{Test}$
\ENSURE $\mathcal{P}(u) = f$ on $\Omega$ and $\mathcal{B}(u) = g$ on $\partial\Omega$
\STATE Initialize $iter = 0$, $\mathcal{S}_{CL}$, $\mathcal{S}_{BC}$, and $\mathcal{S}_{PL}^{NN} = \mathcal{S}_{PL}^{GP} = \emptyset$
% AUTHOR CHECK: With iter starting at 0, this loop permits I_max + 1 passes. Confirm whether I_max is a final index or a count; the original stopping rule is retained.
\WHILE{$iter \le I_{\max}$}
\STATE Train $u_{NN}$, $u_{GP}$ for a fixed number of epochs on $\mathcal{S}_{CL}$, $\mathcal{S}_{BC}$ and $\mathcal{S}_{PL}^{NN}, \mathcal{S}_{PL}^{GP}$ respectively
\STATE Randomly generate $\mathcal{S}_{Test}^{iter}$
\STATE Find all possible $\bx_i^{Test, iter}\in \mathcal{S}_{Test}^{iter}$, such that
\[
% AUTHOR CHECK: Both distance tests use the GP pseudo-label set. Confirm whether the PINN test should instead use the NN set; the original algorithm is retained.
\abss{\mathcal{P}(u_{NN})(\bx_i^{Test, iter}) - f(\bx_i^{Test, iter})} < \sigma_1 \quad \text{and} \quad \text{dist}(\bx_i^{Test, iter}, \mathcal{S}_{BC}^{\bx} \cup \mathcal{S}_{PL}^{\bx, GP}) < \sigma_2,
\]
\STATE then $(\bx_i^{Test, iter}, u_{NN}(\bx_i^{Test, iter}))$ is moved to $\mathcal{S}_{PL}^{GP}$;
\STATE Again find all possible $\bx_i^{Test, iter}\in \mathcal{S}_{Test}^{iter}$, such that
\[
\text{Var}(u_{GP})(\bx_i^{Test, iter}) < \sigma_1 \quad \text{and} \quad \text{dist}(\bx_i^{Test, iter}, \mathcal{S}_{BC}^{\bx} \cup \mathcal{S}_{PL}^{\bx, GP}) < \sigma_2,
\]
\STATE then $(\bx_i^{Test, iter}, u_{GP}(\bx_i^{Test, iter}))$ is moved to $\mathcal{S}_{PL}^{NN}$
\IF{No such $\bx_i^{Test, iter}$ is found for both cases}
  \STATE Break
\ELSE
  \STATE $iter = iter + 1$
\ENDIF
\ENDWHILE
\end{algorithmic}
\end{algorithm}
\section{Numerical Experiments}
We test the self-training and co-training schemes of PINN and PIGP on several important classes of PDEs, i.e., viscous Burgers with small viscosity, linear transport with fast wave speed, Allen--Cahn with sharp transition, and Helmholtz equation with inhomogeneous local structures.  One way to combine PINN and PIGP methods is the co-training described above. However, it is also possible to combine machine learning methods based on Neural Networks (NNs) and Gaussian Processes (GPs) within the same models, which leads to hybrid PINN-PIGP methods. For certain classes of models (especially linear models), GPs provide an exact solution which is extremely computationally efficient (because it might have closed-form solutions).  However, NNs can be used to learn and approximate models which have significant non-linearities. Our aim is also to develop methods that handle the tractable parts of the model with the closed-form GPs (similarly to Rao-Blackwellization in the state-space context \citep{Sarkka:2013}) and the non-linear parts with approximate GPs or fully non-linear NNs. This way we can combine the advantages of the GP and NN methods in solving general physics-informed machine learning problems. 
\subsection{Benchmarking: Viscous Burgers with Small Viscosity}
We first test the self-training scheme for PINN on an important class of parabolic PDEs, namely the Viscous Burgers (vBurgers) equation, which is to show that by repeatedly propagating the pseudo-labeled points, we can improve the solution accuracy even with small viscosity.  The vBurgers PDE is shown as follows,
\begin{equation}\label{eq:vBurgers}
\begin{aligned}
u_t + (f(u))_x &= \nu u_{xx}, \quad f(u) = \frac{u^2}{2}, \quad (t, x) \in (0, T)\times(a, b), \\
u(0, x) &= u_0(x), \quad x \in [a, b], \\
u(t, a) &= u_a, \quad u(t, b) = u_b, \quad t \in [0, T], \\
\end{aligned}
\end{equation}
where $\nu > 0, T > 0, a < b, u_a, u_b$ are prescribed constants.  Here $\nu$ is also known as the viscosity constant, and the smaller $\nu$ gets, the sharper transitions a solution can have.  Vanilla PINN can handle $\nu = \frac{0.01}{\pi}$ without further modification. We demonstrate the effectiveness of our self-training scheme by testing on the vBurgers on a series of small training epochs with the following parameters
\[
u_0(x) = -\sin(\pi x), \quad u_a = 0, \quad u_b = 0, \quad T = 1, \quad a = -1, \quad b = 1, \quad \nu = \frac{0.004}{\pi}.
\]
We train a PINN solution using the following parameters in Table \ref{tab:pinn_self_setup_vBurgers}.
\begin{table}[!h]
\centering
\begin{tabular}{c | c | c | c | c}
\# Co. Points & \# IC Points & \# L/R BC Points &\# Layers & \# Neurons \\
\hline
$10^4$        & $100$        & $200$           & $8$       & $20$ \\
\end{tabular}
\caption{PINN Parameters}
\label{tab:pinn_self_setup_vBurgers}
\end{table}
It trains with $10k$ Adam steps with a learning rate of $5 \cdot 10^{-3}$ and the fixed loss weights are $(0.1, 1, 0.5, 0.5)$ for PDE residual loss, IC loss, left-boundary loss, and right-boundary loss, respectively; when there are pseudo-labeled points, the corresponding loss weight will be set at $0.05$.  We set the thresholds for total loss at $10^{-7}$, residual loss at $5 \cdot 10^{-6}$, the distance (close to any true data points) at $10^{-2}$, and randomly generate $1000$ test points to become pseudo-labeled points.

\begin{figure}[!h]
\centering
\includegraphics[width=0.45\textwidth]{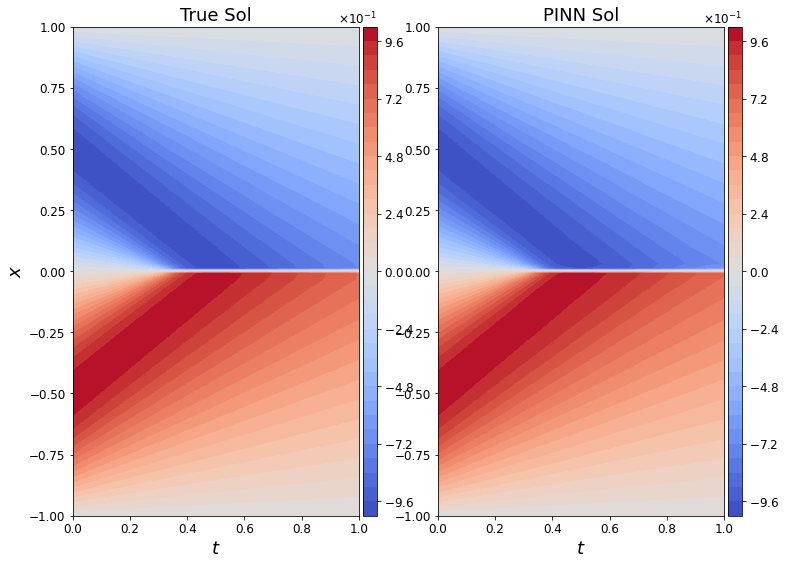}
\,
\includegraphics[width=0.4\textwidth]{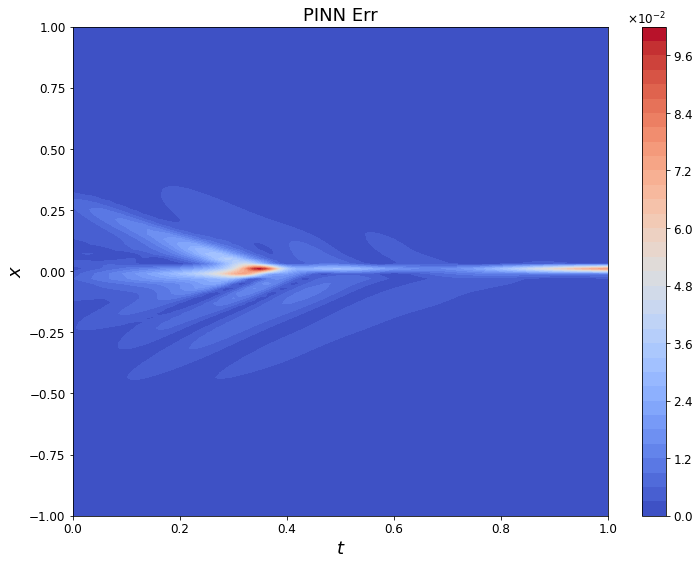}
\caption{No self-training, Rel $L_2$-Err = $9.3 \cdot 10^{-3}$, Rel $L_{\infty}$-Err = $7.4 \cdot 10^{-3}$.}
\label{fig:pinn_self_0train_vBurgers}
\end{figure}

\begin{figure}[!h]
\centering
\includegraphics[width=0.45\textwidth]{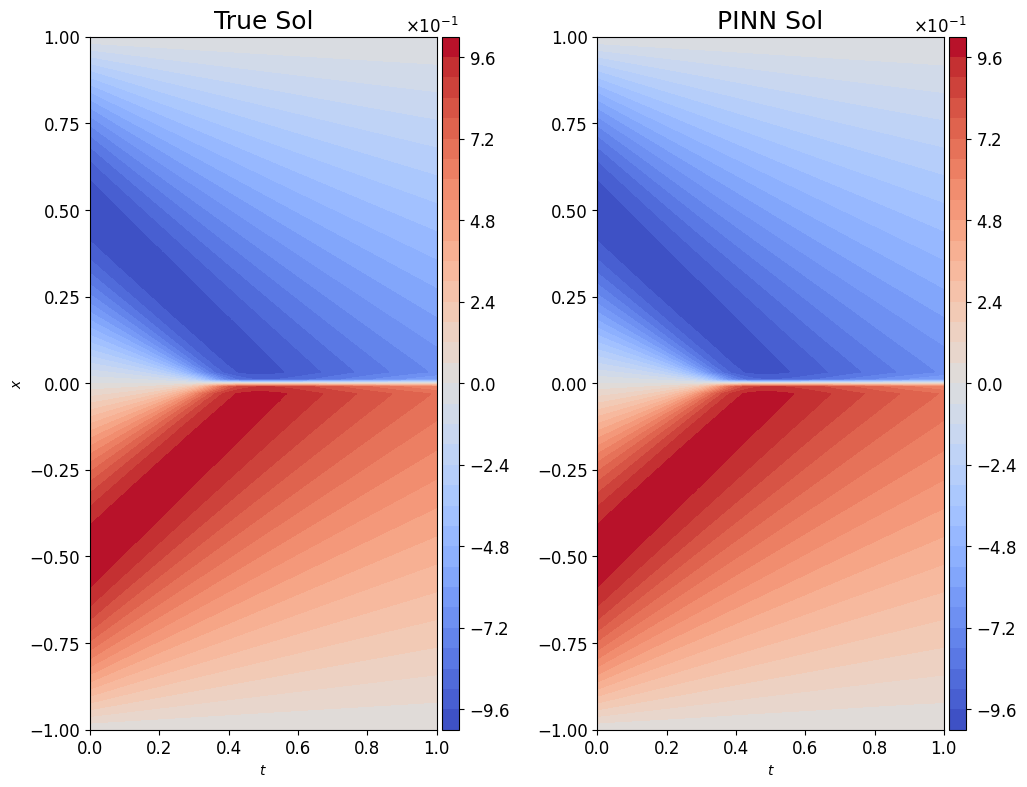}
\,
\includegraphics[width=0.4\textwidth]{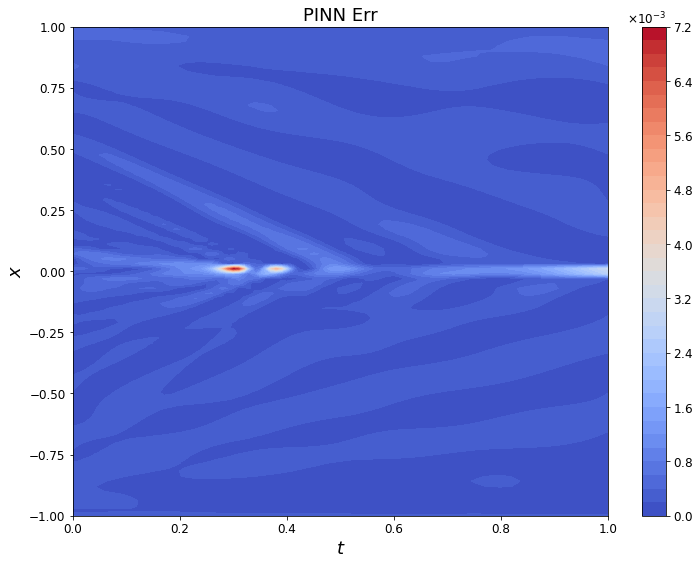}
\caption{$3$ self-training iterations, Rel $L_2$-Err = $1.8 \cdot 10^{-4}$, Rel $L_{\infty}$-Err = $1.5 \cdot 10^{-3}$.}
\label{fig:pinn_self_5train_vBurgers}
\end{figure}

\begin{figure}[!h]
\centering
\includegraphics[width=0.5\textwidth]{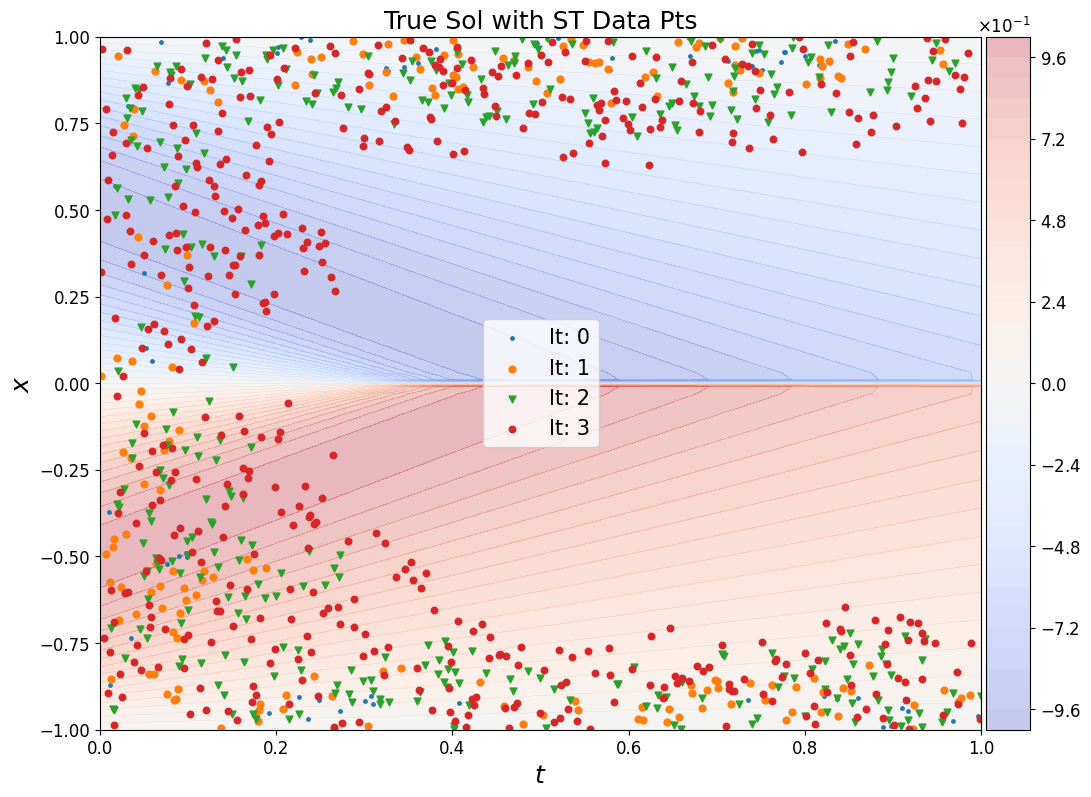}
\caption{Propagation of Labeled Points in a self-trained PINN solving vBurgers.}
\label{fig:pinn_self_pt_prop_vBurgers}
\end{figure}

% AUTHOR CHECK: Burgers iteration count: the prose says 4, the caption says 3, and the plot legend spans 0--3. Confirm the counting convention against the experiment.
\textbf{Test Results}: We perform $4$ self-training iterations for a PINN solving the viscous Burgers PDE.  Figure \ref{fig:pinn_self_0train_vBurgers} shows the result from PINN before the self-training. Figure \ref{fig:pinn_self_5train_vBurgers} shows the result from PINN after $4$ self-training iterations. Figure \ref{fig:pinn_self_pt_prop_vBurgers} shows the propagation of labels in self-training points.

% AUTHOR CHECK: Burgers error values below conflict with the captions (9.3e-3 before and 1.8e-4 after). Original numerical values are retained pending verification.
\textbf{Conclusion}: By repeatedly feeding reliable pseudo data points from a PINN approximation to itself, one can improve the results from PINN significantly (from $9.3 \cdot 10^{-4}$ in $L_2$-Err to $5.4\cdot10^{-4}$ in $L_2$).  Another interesting observation is that the points show roughly the direction of the time evolution (from the IC/BC points inwards and points on IC/BC are labeled), since the PINN training scheme does not enforce the direction of time in the objective loss, but the self-training scheme can pick up this direction.
\subsection{High Speed Linear Transport Equation}
Next, we test our self-training scheme of using a PINN on a more difficult example, namely the transport equation with a high constant speed given as follows
\begin{equation}\label{eq:Transport}
\begin{aligned}
u_t + cu_x &= 0, \quad (t, x) \in (0, 1)\times(0, 2\pi), \\
u(0, x) &= \sin(x), \quad x \in [0, 2\pi], \\
u(t, 0) &= u(t, 2\pi), \quad t \in [0, 1], \\
\end{aligned}
\end{equation}
As indicated in \cite{krishnapriyan2021characterizing}, when $c\ge 30$,  a normal PINN without any additional training techniques will fail to capture the fast transition of phases.  We demonstrate the effectiveness of our self-training scheme using $c = 30$.  We train a PINN solution using the following parameters in Table \ref{tab:pinn_self_setup_Transport}.
\begin{table}[!h]
\centering
\begin{tabular}{c | c | c | c | c}
\# Co. Points & \# IC Points & \# L/R BC Points &\# Layers & \# Neurons \\
\hline
$10^4$        & $100$        & $200$           & $8$       & $20$ \\
\end{tabular}
\caption{PINN Parameters}
\label{tab:pinn_self_setup_Transport}
\end{table}
It trains with $5k$ Adam steps with a learning rate of $5 \cdot 10^{-3}$ and the fixed loss weights are $(0.1, 1, 0.5)$ for PDE residual loss, IC loss, and periodic boundary loss, respectively; when there are pseudo-labeled points, the corresponding loss weight will be set at $0.1$.  We set the thresholds for total loss at $10^{-5}$, residual loss at $10^{-5}$, and the distance (close to any true data points) at $10^{-2}$.

\begin{figure}[!h]
\centering
\includegraphics[width=0.45\textwidth]{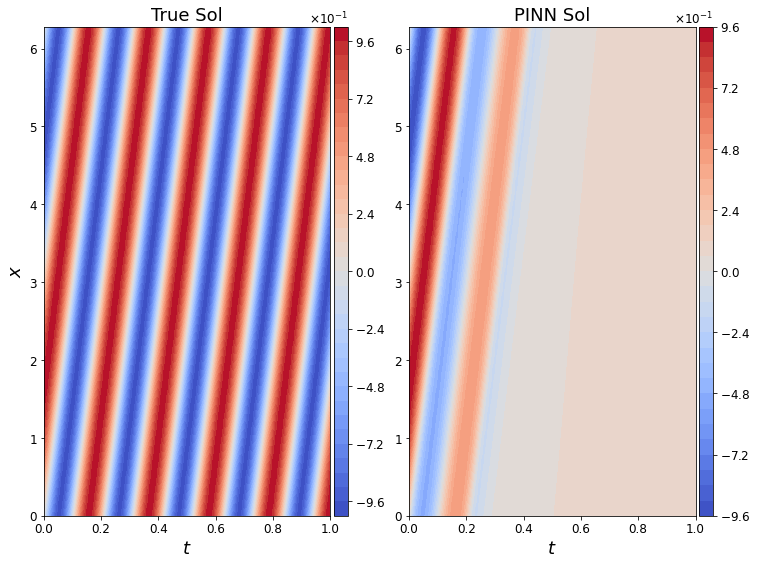}
\,
\includegraphics[width=0.4\textwidth]{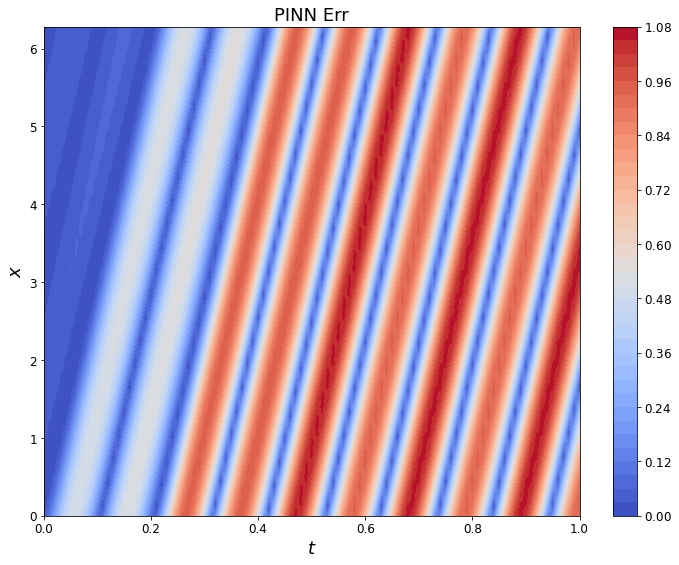}
\caption{No self-training, Rel $L_2$-Err = $6.7 \cdot 10^{-1}$, Rel $L_{\infty}$-Err = $1.0$.}
\label{fig:pinn_self_0train_Transport}
\end{figure}

\begin{figure}[!h]
\centering
\includegraphics[width=0.45\textwidth]{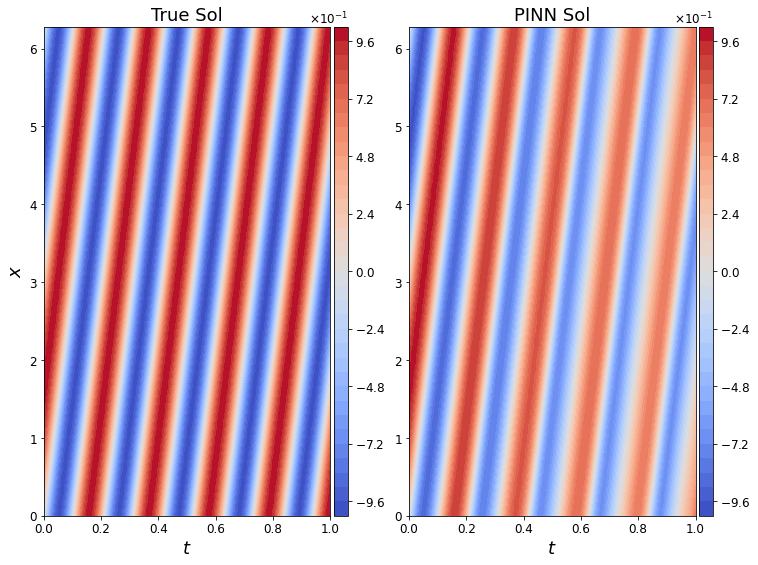}
\,
\includegraphics[width=0.4\textwidth]{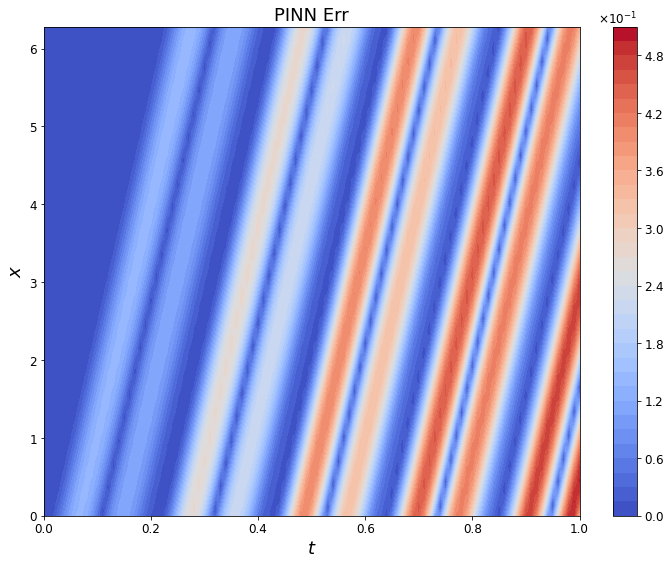}
\caption{$5$ self-training iterations, Rel $L_2$-Err = $4.5 \cdot 10^{-3}$, Rel $L_{\infty}$-Err = $6.8 \cdot 10^{-3}$.}
\label{fig:pinn_self_5train_Transport}
\end{figure}

\begin{figure}[!h]
\centering
\includegraphics[width=0.5\textwidth]{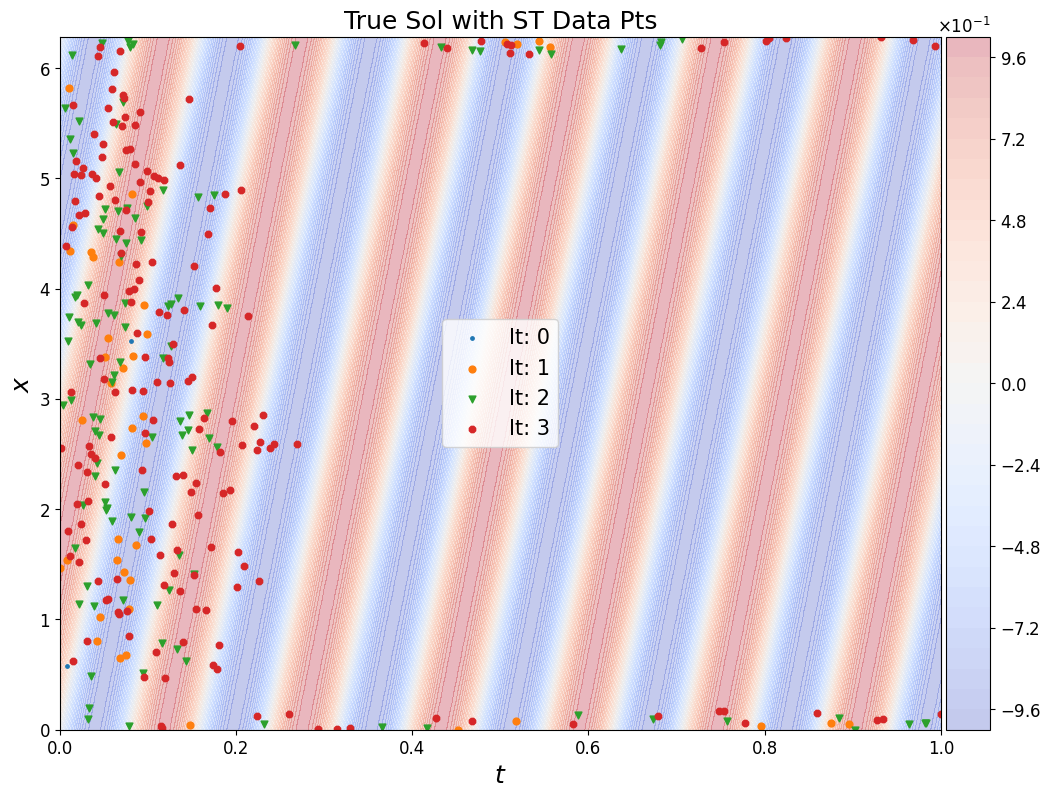}
\caption{Propagation of Labeled Points in a self-trained PINN solving the linear transport PDE.}
\label{fig:pinn_self_pt_prop_Transport}
\end{figure}

\textbf{Test Results}: We perform $4$ self-training iterations for a PINN solving the transport equation with constant speed.  Figure \ref{fig:pinn_self_0train_Transport} shows the result from PINN before the self-training.  Figure \ref{fig:pinn_self_5train_Transport} shows the result from PINN after $4$ self-training iterations. Figure \ref{fig:pinn_self_pt_prop_Transport} shows the propagation of labels in self-training points.

\textbf{Conclusion}: Our self-training scheme can even help a vanilla PINN capture the fast transition of phases by just repeatedly labeling the test points as pseudo-labeled points.  The objective loss in this self-training scheme does not enforce the direction of time flow; however, the selected pseudo-labeled points show some degree of the direction of time.
\subsection{Allen-Cahn Equation: PINN with Time Marching and PINN with GPR}
\begin{figure}[!h]
\centering
\includegraphics[width=0.45\textwidth]{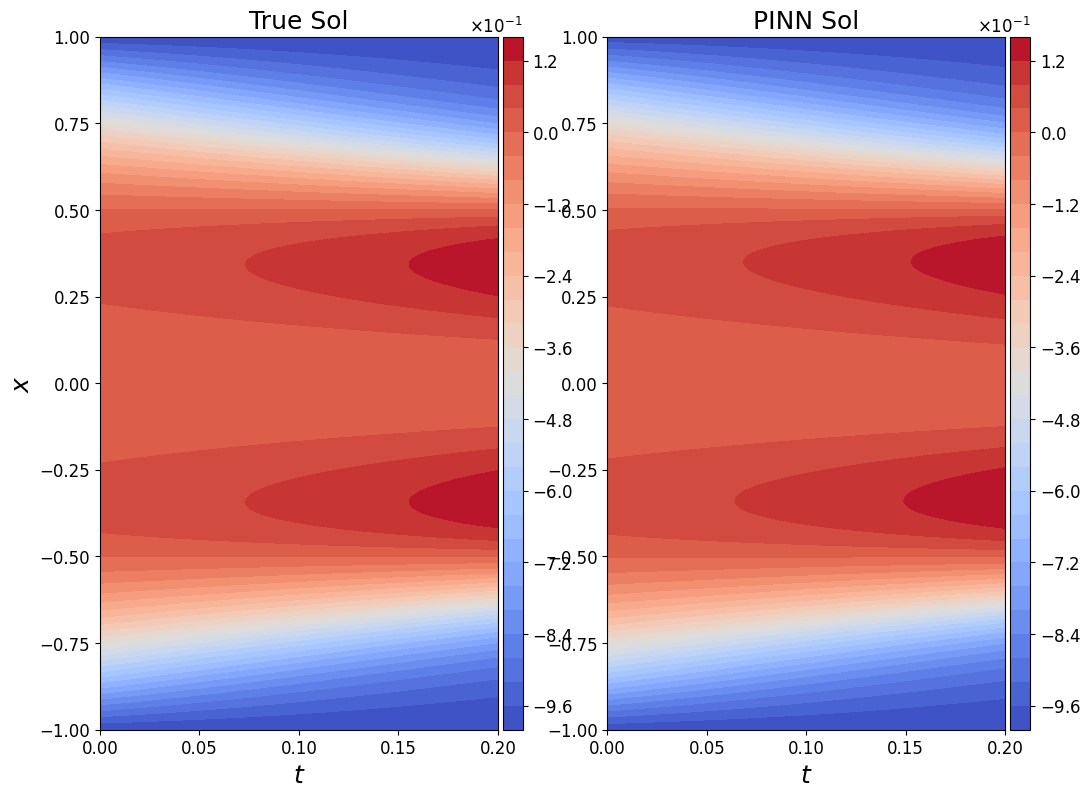}
\,
\includegraphics[width=0.4\textwidth]{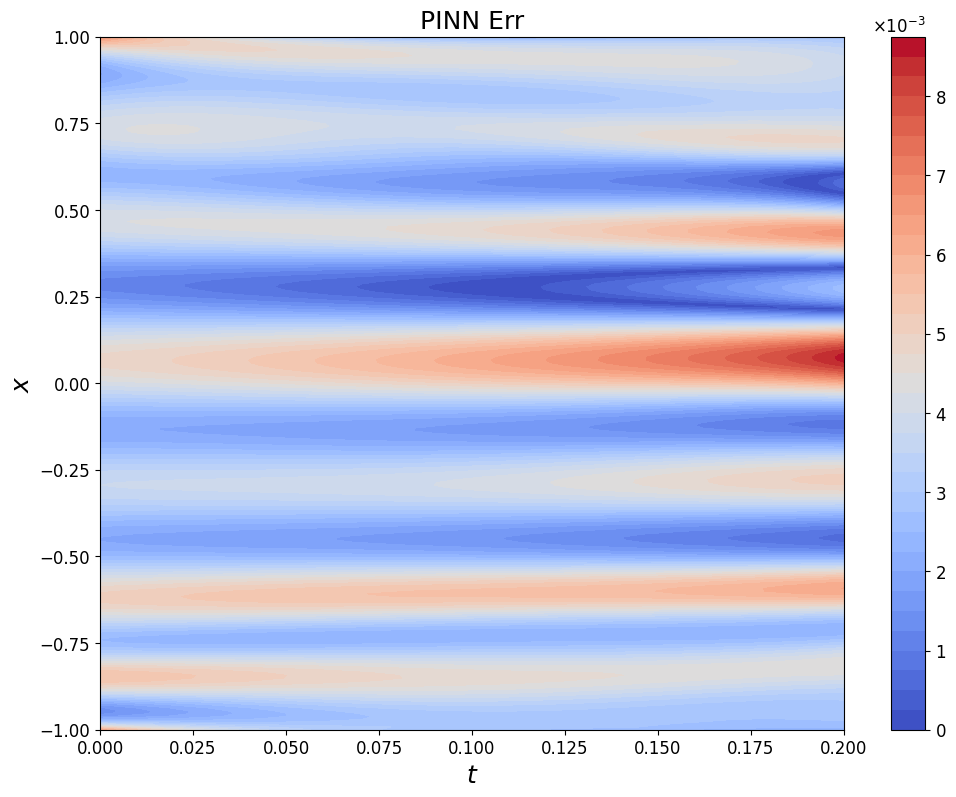}
\caption{No self-training, $L_2$-Err = $3.1 \cdot 10^{-3}$ and $L_{\infty}$-Err = $4.6 \cdot 10^{-3}$.}
\label{fig:pinn_self_0train_AC_TM}
\end{figure}

\begin{figure}[!h]
\centering
\includegraphics[width=0.45\textwidth]{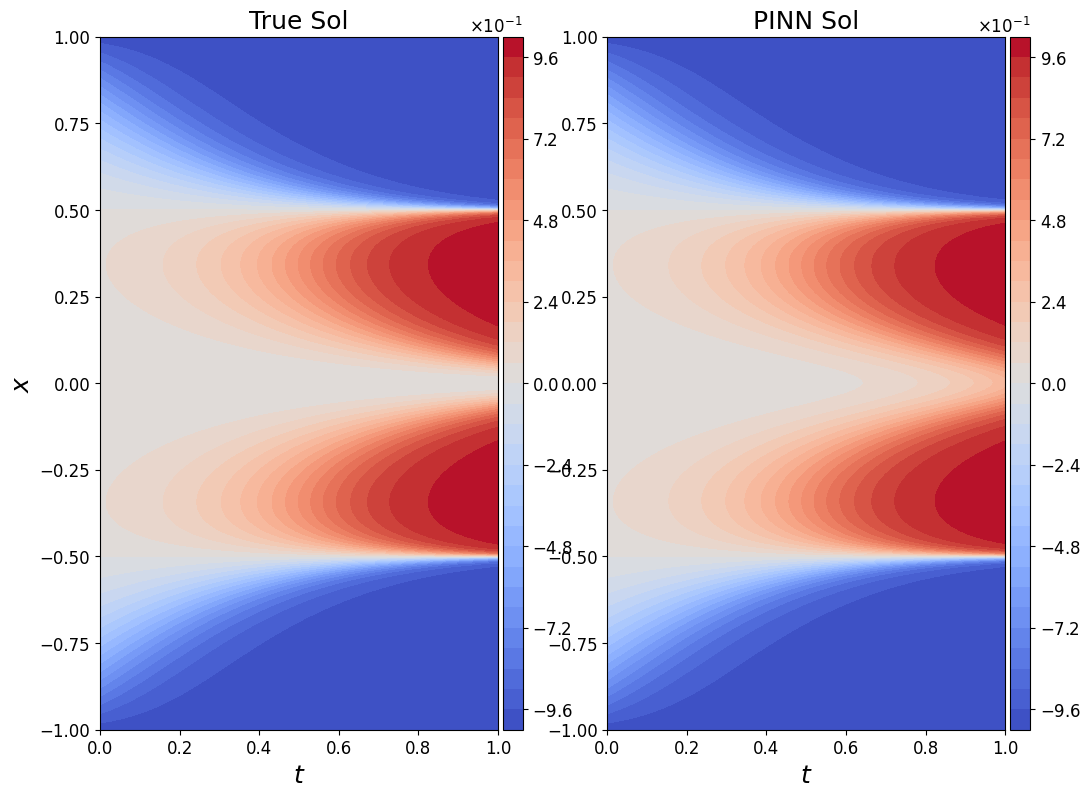}
\,
\includegraphics[width=0.4\textwidth]{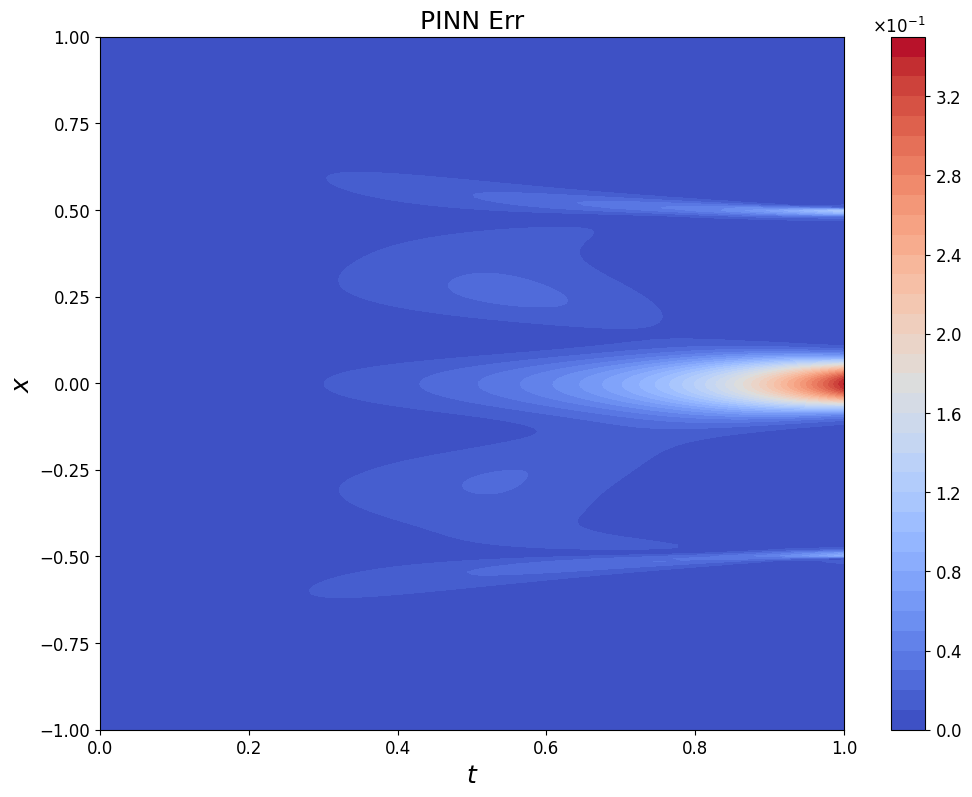}
\caption{$5$ self-training iterations, $L_2$-Err = $2.1 \cdot 10^{-3}$ and $L_{\infty}$-Err = $6.3 \cdot 10^{-3}$.}
\label{fig:pinn_self_5train_AC_TM}
\end{figure}

\begin{figure}[!h]
\centering
\includegraphics[width=0.5\textwidth]{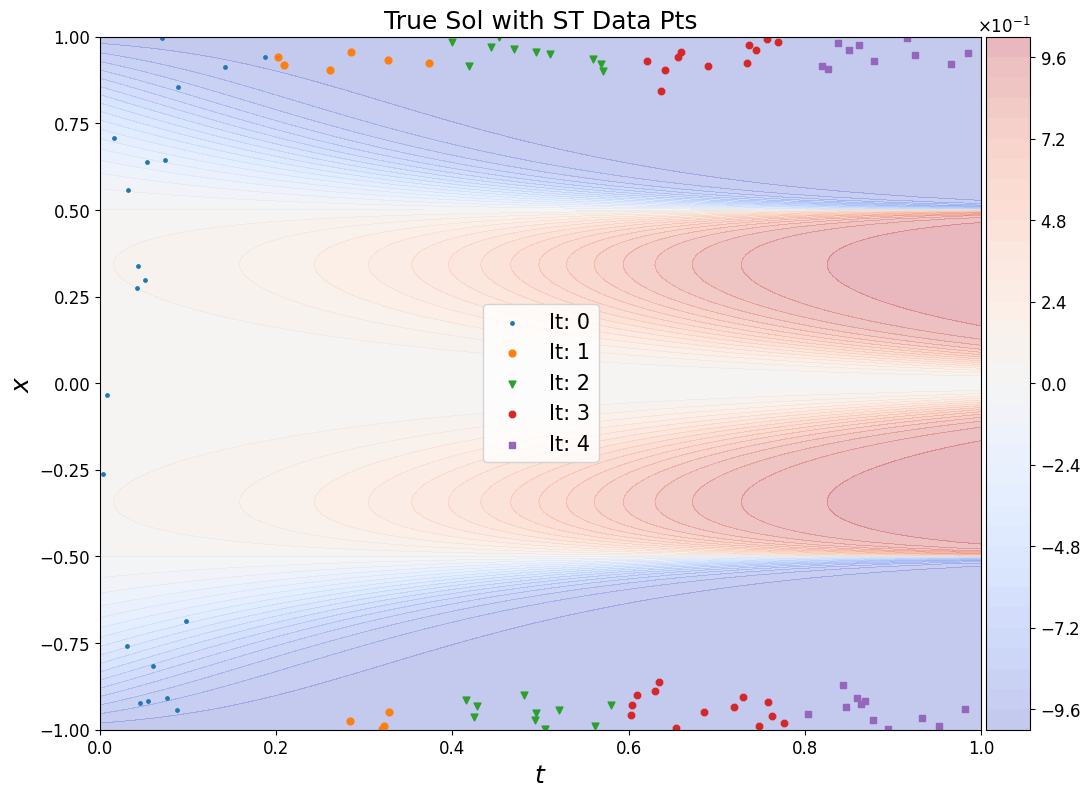}
\caption{Propagation of Labeled Points in a self-trained PINN solving the Allen--Cahn equation with time marching.}
\label{fig:pinn_self_pt_prop_AC_TM}
\end{figure}

The Allen-Cahn PDE is a reaction-diffusion PDE commonly used in phase-field models, which is used to model the phase separation process in the microstructure evolution of metallic alloys~\citep{chen2002phase}.  The Allen-Cahn PDE we consider is given as follows,
\begin{equation}\label{eq:AC_pde}
\begin{aligned}
u_t - 10^{-4}u_{xx} + 5u(u^2 - 1)&= 0, \quad (t, x) \in (0, 1)\times(-1, 1), \\
u(0, x) &= x^2\cos(\pi x), \quad x \in [-1, 1], \\
u(t, -1) &= u(t, 1), \quad t \in [0, 1], \\
\end{aligned}
\end{equation}
We can combine time marching with self-training to improve the learning results from our self-training scheme.  However, we do not require updating the initial condition at each later time as we can obtain pseudo-labeled points as data points (roughly like an initial condition).  We will break the time interval into $5$ equal pieces, i.e. $t_0 = 0$, $t_1 = 0.2$, $t_2 = 0.4$, $t_3 = 0.6$, $t_4 = 0.8$ and $t_5 = 1.0$.  We then train on increasingly larger intervals, i.e. from $[t_{0}, t_{iter}]$ to $[t_0, t_{iter + 1}]$.  Figure \ref{fig:pinn_self_0train_AC_TM} shows the results from PINN before self-training but with a smaller time interval.  Figure \ref{fig:pinn_self_pt_prop_AC_TM} shows the point configuration after $4$ self-training iterations. Figure \ref{fig:pinn_self_5train_AC_TM} shows the result from PINN after $4$ self-training iterations with time marching.
Figure \ref{fig:pinn_self_pt_prop_AC_TM} shows the propagation of labels during self-training.

\begin{figure}[!h]
\centering
\includegraphics[width=0.45\textwidth]{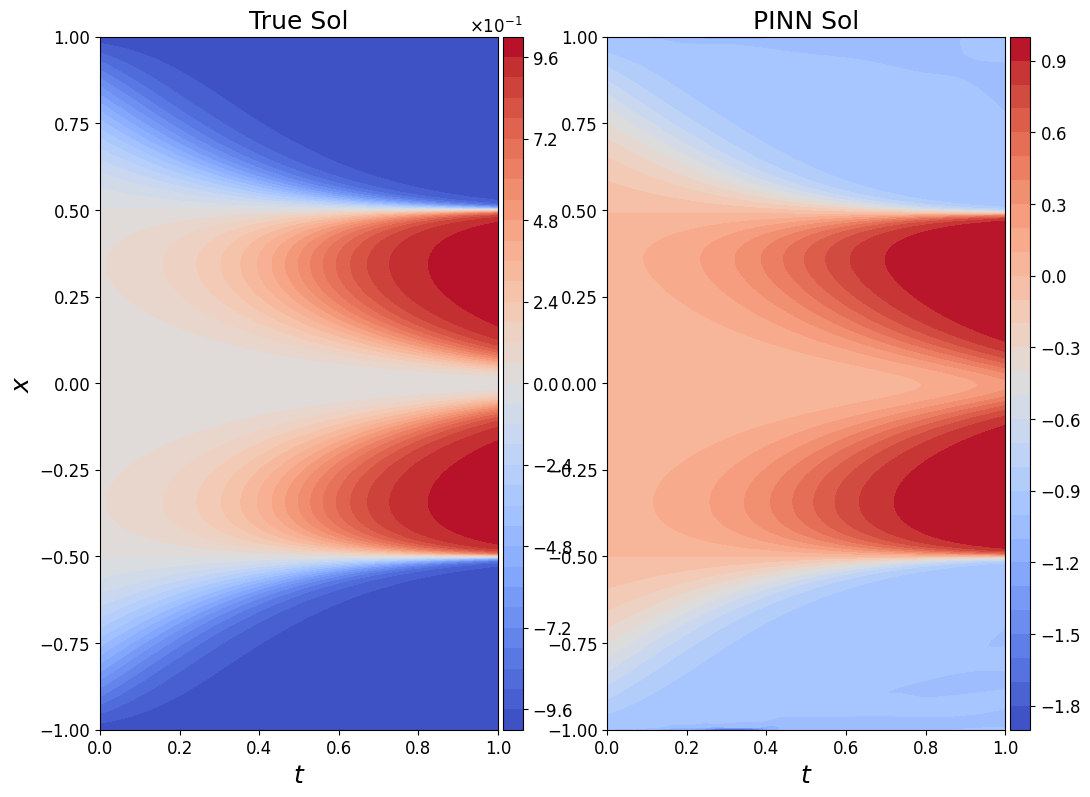}
\,
\includegraphics[width=0.4\textwidth]{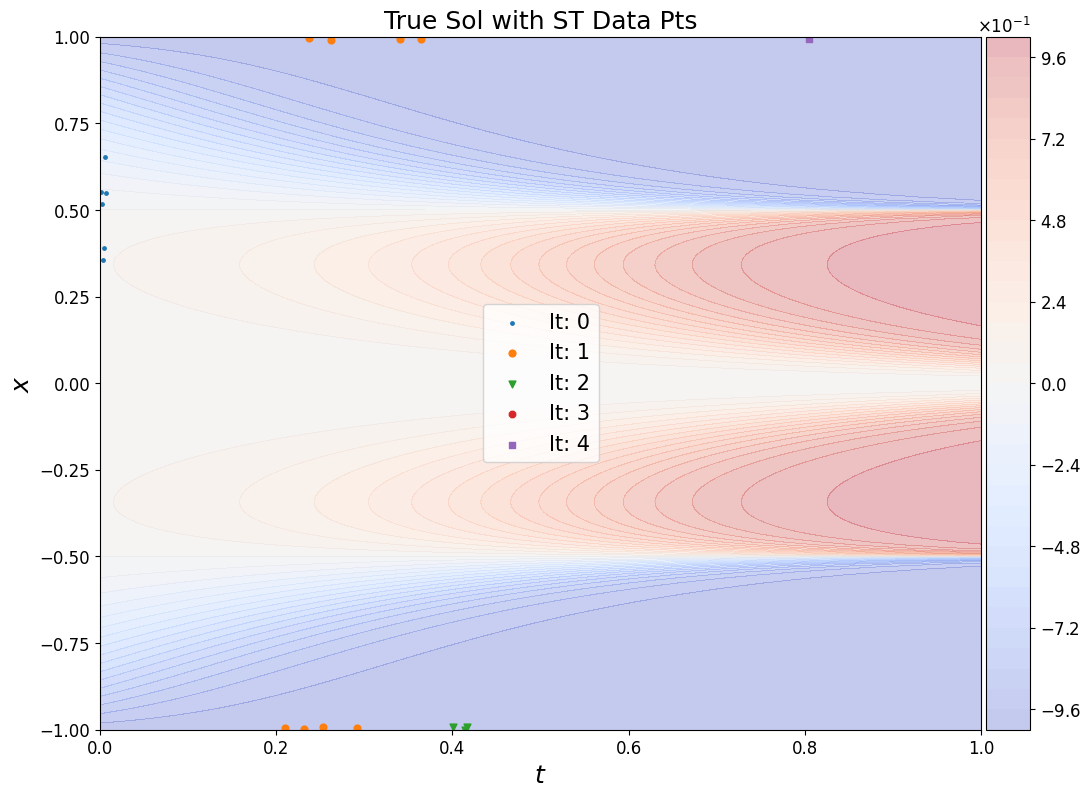}
\caption{PINN Self-Train with GPR, $L_2$-Err $= 7.7 \cdot 10^{-2}$.}
\label{fig:pinn_gpr_self_train}
\end{figure}

Another possible combination for the self-training scheme is to combine a PINN regressor with a Gaussian Process regressor.  Instead of only adding the pseudo-labeled points into the PINN regressor, it also adds them to the GPR and uses cross-referenced criteria (low PDE residual from PINN and low variance from GPR) to determine which points qualify as pseudo-labeled points. The results are shown in Figure \ref{fig:pinn_gpr_self_train}.

\textbf{Conclusion}: Our self-training scheme, when combined with time marching, has achieved a much higher accuracy than the normal self-training scheme.  This time, the pseudo-labeled points concentrated around the initial condition within the first block, and around the two periodic boundaries.
\subsection{Helmholtz Equation (Co-training)}
We notice that one of the many possible failure modes for a vanilla PINN is the inability to capture high-frequency components of a solution.  % AUTHOR CHECK: The incomplete phrase about the PIGP kernel has been repaired as not requiring an optimal kernel; confirm that this is the intended claim.
Hence, we test the co-training scheme of using a PIGP to train a PINN to solve elliptic PDEs in order to overcome spectral bias in the PINN solution, without requiring an optimal choice of the PIGP kernel.  We look at an example of a linear elliptic PDE (non-linear cases can be considered too), namely the Helmholtz equation, as follows
\[
\begin{aligned}
\Delta u(x, y) + k^2 u(x, y) &= f(x, y), \quad (x, y) \in (a, b)\times(c, d), \\
u(a, y) &= u_a(y), \quad y \in [c, d], \\
u(b, y) &= u_b(y), \quad y \in [c, d], \\
u(x, c) &= u_c(x), \quad x \in [a, b], \\
u(x, d) &= u_d(x), \quad x \in [a, b], 
\end{aligned}
\]
We choose the true solution as $u_*(x, y) = \sin(\pi x)\sin(4\pi y)$, and obtain the forcing $f$ and the four boundary conditions accordingly.  We train a PIGP solution with the parameters given by Table \ref{tab:pigp_setup_Helmholtz}.
\begin{table}[!h]
\centering
\begin{tabular}{c | c  }
\# Co. Points & \# BC Points \\
\hline
$1024$        & $132$        \\
\end{tabular}
\caption{PIGP Parameters}
\label{tab:pigp_setup_Helmholtz}
\end{table}
The kernel is chosen as a homogeneous Gaussian, i.e.
\[
K(\bx, \bx') = \exp\!\left(-\frac{1}{2\sigma^2}\sum_{i = 1}^d (x_i - x_i')^2\right), \quad \bx, \bx' \in \R^d.
\]
\begin{figure}[!h]
\centering
\includegraphics[width=0.7\textwidth]{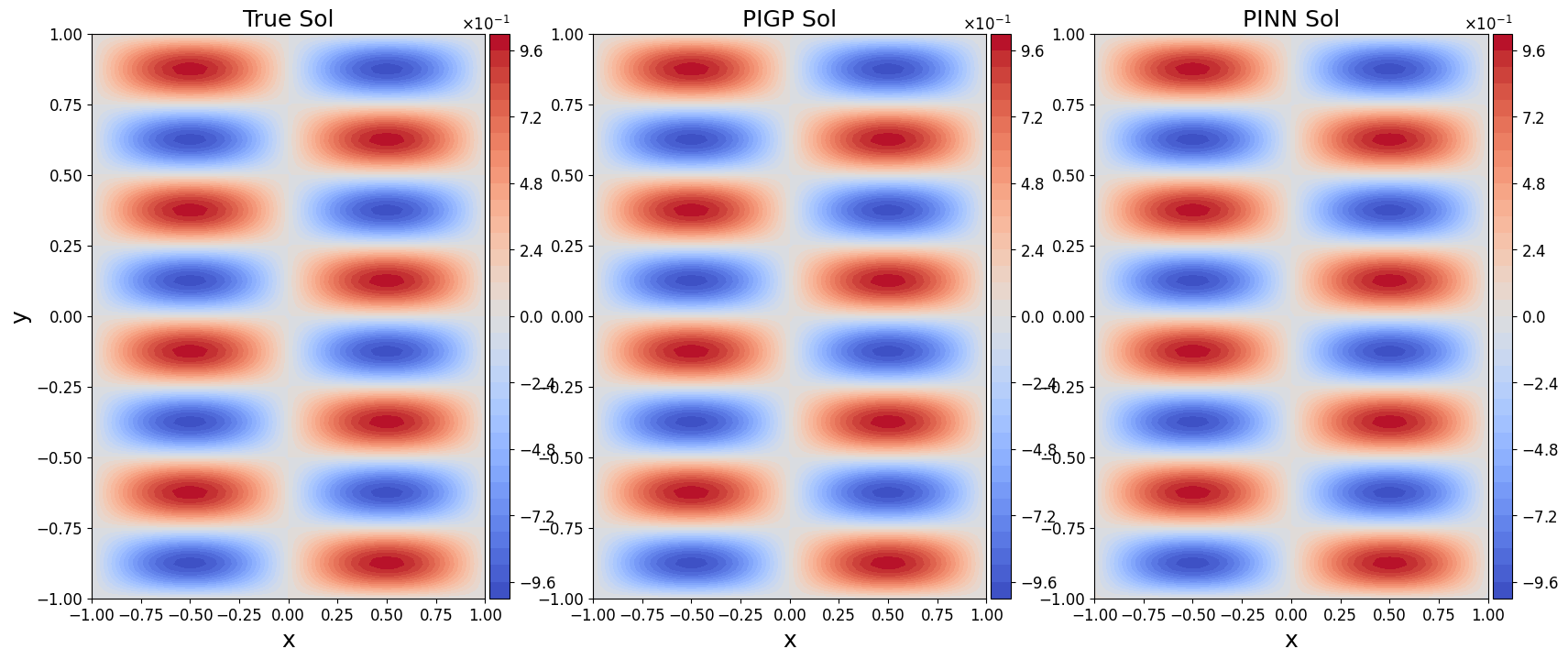}
\caption{PINN and PIGP Co-training, PIGP $L_2$-Err $= 5.7 \cdot 10^{-5}$, PINN $L_2$-Err $= 2.3 \cdot 10^{-3}$}
\label{fig:pinn_0train_Helmholtz}
\end{figure}

We choose the parametric Gaussian kernel with $\sigma = M^{-0.25}$, where $M$ is the total number of points (collocation points and BC points). Such a parameter is shown to be able to handle various kinds of elliptic PDEs (including nonlinear PDEs) in \cite{chen2021solving}.  Then we set up a PINN training using the parameters given by Table \ref{tab:pinn_setup_Helmholtz}.
\begin{table}[!h]
\centering
\begin{tabular}{c | c | c | c }
\# Co. Points & Points per Boundary &\# Layers & \# Neurons \\
\hline
$10^4$        & $100$          & $7$       & $20$ \\
\end{tabular}
\caption{PINN Parameters}
\label{tab:pinn_setup_Helmholtz}
\end{table}
\begin{figure}[!h]
\centering
\includegraphics[width=0.7\textwidth]{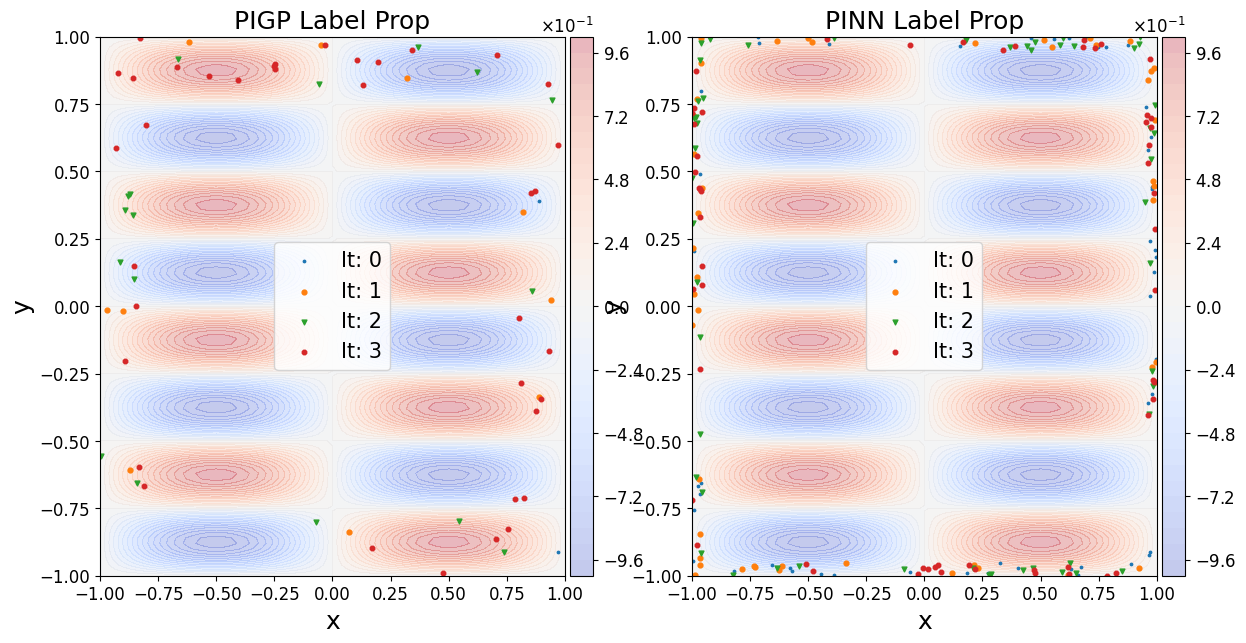}
\caption{Propagation of Labels in PIGP-PINN co-training for the Helmholtz equation.}
\label{fig:pinn_pigp_cotrain_pt}
\end{figure}
It trains with $5k$ Adam steps with a learning rate of $5 \cdot 10^{-3}$ and the fixed loss weights are $(1, 1, 1, 1, 1)$. If pseudo-labeled data are present, they have a loss weight of $0.1$.

\textbf{Test Results}:  We perform $5$ co-training iterations by using PIGP and PINN as trainers for each other for solving the Helmholtz PDE.  Figure \ref{fig:pinn_0train_Helmholtz} shows both the PINN and PIGP results after co-training. Figure \ref{fig:pinn_pigp_cotrain_pt} shows the propagation of labels in co-training points.
\textbf{Conclusion}: By feeding the pseudo-labeled points each other by solutions from PINN and PIGP simultaneously, the PINN and PIGP estimators are able to capture some of the high-frequency components in its solution.  Training took $5{,}000 \times 6 = 30{,}000$ Adam iterations and achieved an $L_2$ error of $10^{-2}$. This required fewer iterations than reported in \cite{wang2020understanding}, but slightly more than reported in \cite{mcclenny2020self}.
\section{Conclusion}
We presented a semi-supervised label-propagation framework for training physics-informed neural networks (PINNs) and physics-informed Gaussian processes (PIGPs). By progressively assigning pseudo-labels to collocation points with sufficiently reliable predictions, self-training and co-training provide a mechanism for propagating information from initial and boundary data into the computational domain. Our numerical experiments demonstrate that these strategies can improve prediction accuracy across several PDE benchmarks, including problems that are challenging for standard PINN training.

A central feature of the framework is the co-training of structurally independent predictors with complementary properties. PINNs and PIGPs exchange pseudo-labels using model-specific confidence criteria, while the GP component provides posterior uncertainty estimates that help guide label selection. For nonlinear problems, the PINN–GPR combination offers an additional way to incorporate probabilistic predictions into the training process. These strategies can also be combined with time marching and other training enhancements. Future work will investigate the sensitivity to confidence thresholds, the accumulation of pseudo-label errors, and the calibration of uncertainty estimates, with the aim of establishing when and why co-training yields reliable improvements.
%
%\subsubsection*{Broader Impact Statement}
%In this optional section, TMLR encourages authors to discuss possible repercussions of their work,
%notably any potential negative impact that a user of this research should be aware of. 
%Authors should consult the TMLR Ethics Guidelines available on the TMLR website
%for guidance on how to approach this subject.
%
%\subsubsection*{Author Contributions}
%If you'd like to, you may include a section for author contributions as is done
%in many journals. This is optional and at the discretion of the authors. Only add
%this information once your submission is accepted and deanonymized. 
%
\subsubsection*{Acknowledgments}
Zhong and Braga-Neto are supported by NSF-AoF grant $\#2225507$.
\bibliography{cotrain}
\bibliographystyle{tmlr}
%
%\appendix
%\section{Appendix}
%You may include other additional sections here.
%
\end{document}